%% file: main.tex
\documentclass[conference]{IEEEtran}
\IEEEoverridecommandlockouts
\usepackage{cite}
\usepackage{amsmath,amssymb,amsfonts}
\usepackage{graphicx}
\usepackage{kotex}
\usepackage{textcomp}
\usepackage{xcolor}
\usepackage{fvextra}
\usepackage{algorithm}
\usepackage{algpseudocode}
\usepackage{array}  % For column formatting
\usepackage{tabularx}
\usepackage{booktabs}

\def\BibTeX{{\rm B\kern-.05em{\sc i\kern-.025em b}\kern-.08em
    T\kern-.1667em\lower.7ex\hbox{E}\kern-.125emX}}
\begin{document}

\title{TeXBLEU: Automatic Metric for Evaluate LaTeX Format}

\author{
Kyudan Jung$^{1}$, Nam-Joon Kim$^{2}$, Hyun Gon Ryu$^{3}$, Sieun Hyeon$^{2}$, Seung-jun Lee$^{1}$, Hyuk-jae Lee$^{2}$\\
$^{1}$Chung-Ang University, Seoul, Korea \\
$^{2}$Seoul National Univeristy, Seoul, Korea \\
$^{3}$NVIDIA
}

\maketitle

\begin{abstract} LaTeX is suitable for creating specially formatted documents in science, technology, mathematics, and computer science. Although the use of mathematical expressions in LaTeX format along with language models is increasing, there are no proper evaluation matrices to evaluate them. In this study, we propose TeXBLEU, a metric for evaluating mathematical expressions in the LaTeX format built on the n-gram-based BLEU metric widely used in translation tasks. The proposed TeXBLEU consists of a predefined tokenizer trained on the arXiv paper dataset and a fine-tuned embedding model with positional encoding. The TeXBLEU score was calculated by replacing BLUE's modified precision score with the similarity of n-gram-based tokens. TeXBLEU showed improvements of 86\%, 121\%, and 610\% over traditional evaluation metrics, such as BLEU, sacreBLEU, and Rouge, respectively, on the MathBridge dataset with 1,000 data points. The code is available at https://github.com/KyuDan1/TeXBLEU. \end{abstract}


\section{Introduction}

With the rise of large language models (LLMs), both the public and researchers are using them in various applications. One of these tasks is to generate mathematical formulas in the LaTeX format \cite{latexgcl, llmassisted, arooj2020web, blind, manzoor2019alap, sanmiguel2015latexvoice, baker2021editing}. For example, in converting spoken mathematical expressions into LaTeX, as shown by Jung \cite{mathbridge}, a suitable evaluation metric is required. This metric must assess whether the generated LaTeX output correctly represents the intended formula. In such cases, the generated LaTeX expressions may differ from the ground truth in form yet still convey the same mathematical meaning. This is similar to how different natural language expressions have the same meaning. 

LaTeX is a command-based language, much like programming languages such as C or SQL. It also includes special characters. These features make conventional natural language metrics unsuitable for evaluating mathematical LaTeX expressions, as shown in Figure 1 \cite{BLEU,tangledbleu,reiter2018structured, saadany-orasan-2021-bleu,sacreBLUE}.

\begin{figure}[t] \centering \includegraphics[width=0.5 \textwidth, trim=90mm 32mm 90mm 32mm, clip]{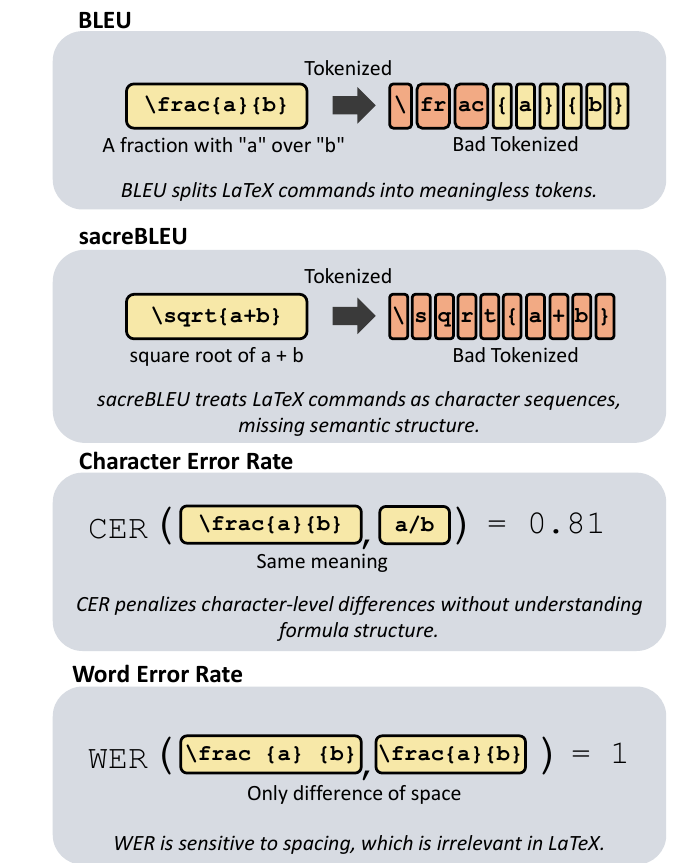} \caption{Limitations of existing metrics in evaluating mathematical expressions in LaTeX format. Both BLEU and sacreBLEU fail to preserve meaning due to tokenizers that are incompatible with LaTeX, resulting in poorly tokenized expressions (highlighted in red). A lower CER generally indicates higher similarity. However, we observed high CER values even when identical LaTeX expressions were input. For WER, minor differences in spacing caused all words to be marked as non-matching, leading to a WER of 1.} \label{fig } \end{figure}

To evaluate LaTeX mathematical expressions effectively, different expressions that compile the same formula should be rated similarly. For instance, using `/' instead of `\textbackslash frac,' or `$\cdot$' instead of `\textbackslash cdot.' Conversely, LaTeX expressions that appear similar but produce different formulas should be distinctly assessed. This occurs owing to variations in the braces or command scopes. Moreover, the spacing in LaTeX is often inconsistent. Therefore, a robust metric should not be affected by the spacing. 

To address these challenges, we propose TeXBLEU, a metric based on BLEU, but adapted for LaTeX expression evaluation. We developed a new tokenizer, trained on arXiv LaTeX papers, and fine-tuned a pre-trained GPT-2 embedding model to handle LaTeX better. Positional encoding for tokens was also considered to accurately capture syntactic relationships. We also preprocess spacing in a consistent manner to avoid giving undue importance.

TeXBLEU was tested on outputs from an LLM that converted spoken mathematical expressions into LaTeX. We assumed that human evaluation is the most reliable metric. Thus, we compared its performance with those of existing metrics and human evaluations \cite{humaneval1,humaneval2,humaneval3}. We focused on how well TeXBLEU’s correlation coefficient aligned with human evaluations compared to other metrics. For BLEU, the average correlation coefficient was 0.38. In contrast, TeXBLEU achieved a much higher value of 0.71. These results show that TeXBLEU closely aligns with human judgment when evaluating the LaTeX output.
% 숫자 쓰기.

%_------------------------Proposed Method-----------------------------
\begin{figure}[t]
    \centering
    \includegraphics[width=0.4 \textwidth, trim=105mm 68mm 105mm 80mm, clip]{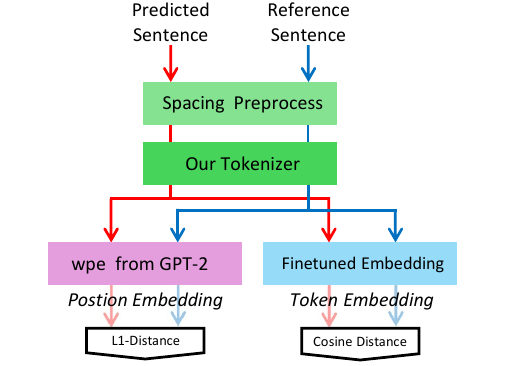}
    \caption{An illustration of the key calculation process of TeXBLEU. First, spacing preprocessing is applied to both the predicted and reference sentences. Then, the sentences are tokenized using a tokenizer built on our LaTeX corpus. Positional encodings are obtained from GPT-2's wpe for each token, and token embeddings are extracted from a fine-tuned GPT-2 embedding model.}
    \label{fig:enter-label}
\end{figure}

\section{Proposed Method}

\subsection{Dataset Collection and Development of a LaTeX-Specific Tokenizer and Embedding Model}
We first created a tokenizer and embedding model based on mathematical expressions in the LaTeX format. To do this, we require a dataset rich in LaTeX-formatted content. The most extensive source of LaTeX-based documents is arXiv papers. We followed the method proposed by Clement to bulk download arXiv paper files \cite{1-a, mathbridge}. This method downloads \texttt{.tex} files of all papers from 2023 using a manifest file containing the metadata of the papers. In total, we collected approximately 172 K \texttt{.tex} files. Using this large corpus of LaTeX data, we created a new tokenizer and embedding model. We built a tokenizer based on byte-pair encoding \cite{BPE, bpe2, bpe3, tokenizer}. Unlike other tokenizers, this is designed to capture LaTeX grammar elements and tokenize them in a way that reflects LaTeX's unique structure. Based on this tokenizer and the arXiv corpus, we fine-tuned a publicly available pretrained GPT-2 embedding model \cite{gpt2}.
\input{results}

\subsection{Token Distance}
\input{algorithm}
When both the reference and predicted sentences were tokenized using our tokenizer and embedding model, we calculated the distance between the tokens, as shown in Figure 2. The token distance \( d(t_1, t_2) \) is defined as  \begin{equation} d(t_1, t_2) = \frac{cosDist(e_1, e_2)^\alpha + tanh(\beta \cdot |p_1 - p_2|)}{2} \end{equation}  where \( t_1, t_2 \) are the tokens, \( e_1, e_2 \) are the token embeddings, \( p_1, p_2 \) are the tokens' positional encoding, and \( \alpha, \beta \) are the hyperparameters. The term `cosDist' refers to the cosine distance, which is defined as 1 minus cosine similarity. This allows us to determine the distance between \( t_1 \) and \( t_2 \). Specifically, the cosine distance is defined as the value obtained by subtracting the cosine similarity from 1. That is,  \begin{equation}     cosDist(e_1, e_2) = 1 - cosSim(e_1, e_2). \end{equation} The reason for using cosine distance in Equation 1 is that, in the case of embedding vectors in natural language processing, the direction of the vectors is often more important than their magnitude, as is well known \cite{embed, embedding}. Taking the power of cosine distance allows similar embeddings to be measured as even more similar and dissimilar embeddings as more distinct. It is already known that applying nonlinear power can adjust the variance. Therefore, we applied a power of \( \alpha \) to the cosine distance.

For positional encoding, the absolute difference in position is important; therefore, we computed L1 distance \cite{position1, position2, position3}. In addition, the hyperbolic tangent function was used because it limits the output within the range of \([-1,1]\), thereby reducing the influence of extreme values while being more sensitive to small differences.  
%-------------------9/8 여기부터 다시보기--------------------------
\subsection{N-gram Similarity}
This section describes the n-gram technique \cite{ngram1, ngram2, ngram3, ngram4}. The n-gram similarity \( \text{sim}_{n}(R, P) \) for a reference sentence \( R \) and a predicted sentence \( P \) is given by  \begin{equation} \text{sim}_{n}(R, P) = 1 - \frac{\sum_{i=1}^{L_n} \sum_{j=1}^n d(r_{ij}, p_{ij})}{L_n \cdot n} \end{equation}  where \( n \) is the length of the \( n \)-gram and \( L_n \) is the number of n-grams. where \( d \) denotes the token distance function.

Specifically, as shown in Algorithm 1, TeXBLEU was calculated using the BLEU n-gram technique. First, in a tokenized sentence, \(n\) tokens are grouped to form \(L_n\) n-grams. For each n-gram, the token distance was calculated for \(n\) tokens. This process was repeated using a nested for-loop to sum up the distances. Dividing this by \(L_n\) and \(n\) yields the average n-gram distance. Finally, to shift the concept to a similarity score rather than distance, n-gram similarity was defined as the n-gram distance subtracted from 1, similar to how it was performed in Equation 2.

\subsection{TeXBLEU}
Based on Algorithm 1 and the explanation above, TeXBLEU is finally equal to the following expression:

\begin{equation}
\text{TeXBLEU} = \exp\left(\sum_{n=1}^N w_n \log \text{sim}_n(R, P)\right)
\end{equation}

After the English sentence is inputted, consistent spacing is applied to LaTeX commands, braces, and other elements, as shown in Algorithm 1 \cite{preprocess1, preprocess2}.

On the other hand, the original BLEU metric employs a brevity penalty (BP). This is intended to penalize cases in which the translation result is excessively short. However, in the LaTeX format, even with differences in sentence length, meaning can remain highly similar. For example, `\textbackslash frac\{\}\{\}' and `/' both represent a fraction, where a single character and nine characters convey a similar meaning. Therefore, the brevity penalty used in the original BLEU was not applied to TeXBLEU. In fact, when the brevity penalty was applied, it resulted in a lower performance compared to TeXBLEU.

\begin{figure}[t]
    \centering
    \includegraphics[width=0.4 \textwidth, trim=95mm 58mm 95mm 58mm, clip]{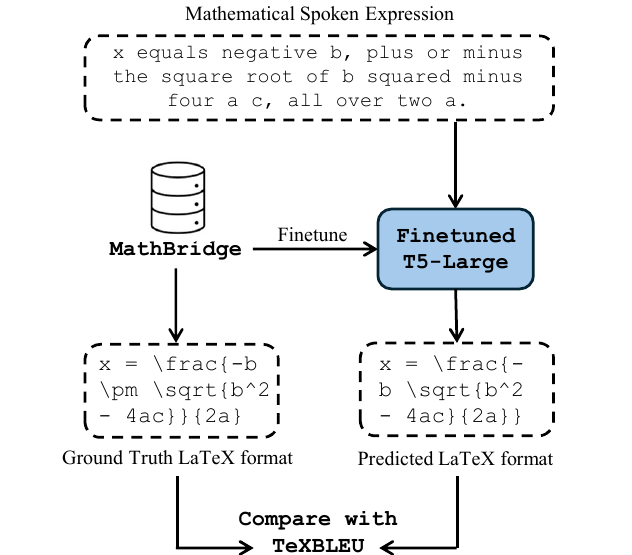}
    \caption{An illustration depicting the main experiment. The T5-Large model, fine-tuned on the MathBridge dataset, is used to input mathematical spoken expressions. The predicted LaTeX format output is then compared to the original ground truth in MathBridge using various metrics.}
    \label{fig:enter-label}
\end{figure}

\begin{figure*}
    \centering
    \includegraphics[width=0.9 \textwidth, trim=42mm 86mm 42mm 87mm, clip]{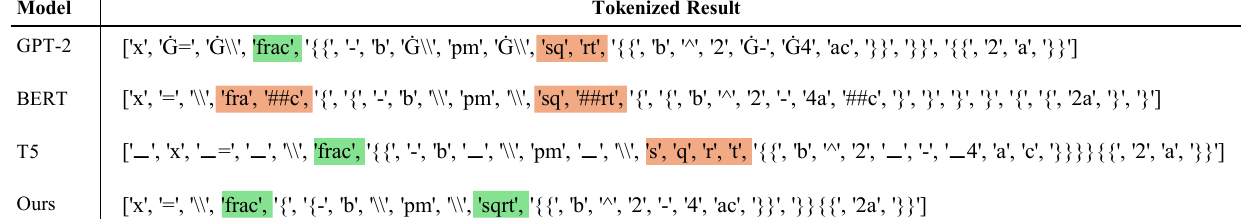}
    \caption{The results of tokenizing the quadratic formula in LaTeX format using various models. The green boxes indicate sections where LaTeX commands were successfully tokenized as complete chunks, while the red boxes represent sections where the LaTeX commands were fragmented, losing their meaning during tokenization. It is evident that the tokenizer we developed using the arXiv paper corpus performs the best in accurately tokenizing LaTeX commands.}
    \label{fig:enter-label}
\end{figure*}

\section{Experiment}
Two experiments were conducted in this study. First, we performed a primary experiment to determine which metric is the most suitable for evaluating the mathematical expressions of the LaTeX format among various metrics. This experiment was conducted by measuring how closely the correlation with human evaluation aligns on the same dataset. Second, we conducted a smaller experiment to test whether the tokenizer we proposed captures LaTeX commands more effectively than other model tokenizers, as mentioned in the paper.

\subsection{Main Experiment}
\subsubsection{Dataset and Setup}
As shown in Figure 3, we used the MathBridge dataset \cite{mathbridge}, which is a publicly available dataset containing English descriptions of mathematical expressions and their corresponding LaTeX formats. Experiments were conducted on a test dataset containing 1,000 data points using the language model fine-tuned on MathBridge by Jung \cite{mathbridge}. The test dataset was extracted separately from the training dataset. Specifically, we input the English descriptions of mathematical expressions into a T5-large model fine-tuned on MathBridge and compared the output LaTeX format with the ground truth LaTeX format. In previous studies, BLEU, sacreBLEU, CER, and WER have been used for evaluation. We evaluated the same data using TeXBLEU.

Among the hyperparameters for calculating the token distance, \( \alpha \) was set to 2 and \( \beta \) was set to 0.1. Positional encoding directly adopts GPT-2's word position embedding (\texttt{wpe}). In addition, the maximum \( n \) value for the n-grams was set to 3.

\subsubsection{Human Evaluation}
The ideal metric is human evaluation. We conducted evaluations on the test dataset using two groups of human evaluators, H1 and H2, each consisting of five members. They were asked to score the predicted LaTeX format compared to the reference LaTeX format on a scale of 1 to 5, where 1 is ``very inaccurate," 2 is ``inaccurate," 3 is ``neutral," 4 is ``accurate," and 5 is ``very accurate." To assess the performance of TeXBLEU, we evaluated the Pearson correlation coefficient and Spearman's rank correlation coefficient between Human Evaluation and each metric. Pearson’s correlation coefficient between H1 and H2 was 0.92.

\subsubsection{Performance Discussion}
We compared the metrics used by Jung \cite{mathbridge} with those used by TeXBLEU. As shown in Table 1, TeXBLEU had the highest correlation with human evaluations compared to all other metrics. Additionally, WER provided significantly less accurate assessments than the other metrics.

By contrast, \cite{similar} proposed various language models as metrics; however, they have the drawback of being slow. TeXBLEU, which is algorithm-based, is considerably faster. When TeXBLEU was run 10 times on two sentences consisting of 20 characters, the average execution time was 0.015 s.

\subsection{Tokenized Result}
Following the Main Experiment, we conducted a smaller experiment to verify the performance of the tokenizer we proposed. arXiv papers in tex format contain various LaTeX commands. When a BPE tokenizer is created using a corpus containing such text, the frequently occurring LaTeX commands are merged into a single token using the tokenizer generation algorithm \cite{sentencepiece}, unlike corpora that contain only general English. We set the vocabulary size to 30 K. As shown in Figure 4, whereas the tokenized results of other well-known language models mostly split the `sqrt' command, our tokenizer successfully tokenized `sqrt' as a single token.

\input{ablation}

\subsection{Ablation study}

In this section, we discuss the ablation study conducted to evaluate the effectiveness of the tokenizer and positional encoding method developed using the arXiv paper dataset. In Table 2, H1 and H2 represent Human Evaluation Groups 1 and 2, respectively, while P and S represent the Pearson and Spearman rank correlation coefficients, respectively. The results showed that applying only the positional encoding method performed better than applying our tokenizer. Furthermore, when both methods were applied, the correlation between human evaluations was up to 2.2 times higher than the baseline.

\section{Conclusion} In this study, we propose TeXBLEU, an automatic metric for evaluating the accuracy of LaTeX formats. We outlined three key requirements that a new metric for LaTeX evaluation should meet and accordingly built a sophisticated metric. We confirmed that, on the MathBridge dataset, TeXBLEU showed a significantly higher correlation with human evaluation metrics than other metrics, thereby demonstrating the superiority of TeXBLEU. One limitation of this study is that there is no metric to verify whether the LaTeX format, when input into a compiler, can generate an error-free equation image. This is a complex issue because different TeX command sets and compilers can result in different compile errors. Future research should address this issue.
%\newpage

\bibliographystyle{ieeetr} 
\bibliography{references}

%\vspace{12pt}

\end{document}

%% file: results.tex
\begin{table*}[ht]
\centering
\small % 조정 가능한 글꼴 크기: \small, \footnotesize, \scriptsize

\caption{Correlation coefficients between the performance of the T5-large model fine-tuned on MathBridge \cite{mathbridge} and human evaluations. H1 refers to human group 1, and H2 refers to human group 2. Pearson and Spearman refer to the Pearson correlation coefficient and Spearman's rank correlation coefficient, respectively.}

\begin{tabularx}{\textwidth}{@{}l|*{10}{>{\centering\arraybackslash}X}@{}}

\toprule
\textbf{Metric Evaluation} &  \multicolumn{6}{c}{\textbf{Metrics}} \\
\cmidrule(lr){2-7}  &  \textbf{BLEU(\textuparrow)} & \textbf{sBLEU(\textuparrow)} & \textbf{Rouge1(\textuparrow)}  & \textbf{CER(\textdownarrow)} & \textbf{WER(\textdownarrow)} & \textbf{TeXBLEU(\textuparrow)} \\
\midrule

score &0.36& 46.8 &0.82 & 0.26 & 0.49 & 0.76  \\
\midrule
H1-Pearson &   0.32 & 0.33 & 0.12 & 0.28 & -0.07 & \textbf{0.71} \\
H1-Spearman &   0.47 & 0.34 & 0.08 & 0.29 & -0.19 & \textbf{0.75} \\
H2-Pearson &   0.30 & 0.31 & 0.09 & 0.21 & 0.01 & \textbf{0.69} \\
H2-Spearman &   0.41 & 0.29 & 0.10 & 0.22 & 0.09 & \textbf{0.70} \\
\midrule
\textbf{Average} & 0.38 & 0.32 & 0.10 & 0.25 & -0.04 & \textbf{0.71} \\

\bottomrule
\end{tabularx}

\end{table*}

%% file: algorithm.tex
\begin{algorithm}
\caption{Computing TexBLEU}
\begin{algorithmic}[1]
\Require Reference text $R$, Prediction text $P$, max n-gram $N$, weights $\{w_1, \ldots, w_N\}$
\Ensure TexBLEU score

\Function{Preprocess}{$R, P$}
    \State $R, P \gets $ coherently spaced $R, P$
    \Return $R, P$
\EndFunction

\Function{TokenDistance}{$t_1, t_2$}
    \State $e_1, p_1 \gets $ embedding and position of $t_1$
    \State $e_2, p_2 \gets $ embedding and position of $t_2$
    \State $d_{emb} \gets \text{cosDist}(e_1, e_2)^\alpha$
    \State $d_{pos} \gets \tanh(\beta \cdot |p_1 - p_2|)$
    
    \Return $\frac{d_{emb} + d_{pos}}{2}$
\EndFunction

\Function{NGramSimilarity}{$R, P, n$}
    \State $R, P \gets \text{Preprocess}(R, P)$
    \State $L_n \gets \min(|R| - n + 1, |P| - n + 1)$
    \State $d_{total} \gets 0$
    \For{$i \gets 1$ to $L_n$}
        \For{$j \gets 1$ to $n$}
            \State $d_{total} 
            \gets d_{total} + \text{TokenDistance}(R[i+j-1], P[i+j-1])$
        \EndFor
    \EndFor
    
    \Return $1 - \frac{d_{total}}{L_n \cdot n}$
\EndFunction

\State $T_R \gets $ tokenize and embed $R$
\State $T_P \gets $ tokenize and embed $P$

\For{$n \gets 1$ to $N$}
    \State $\text{sim}_{n} \gets \text{NGramSimilarity}(T_R, T_P, n)$
\EndFor

\State $\text{TeXBLEU} \gets \exp(\sum_{n=1}^N w_n \log (\text{sim}_{n}(R,P))$

\Return $\text{TeXBLEU}$
\end{algorithmic}
\end{algorithm}

%% file: ablation.tex
\begin{table} [t]
\caption{Ablation study results. P and S represent the Pearson correlation coefficient and Spearman rank correlation coefficient, respectively.}
    \centering
    \begin{tabular}{l|cc|cc}
    \toprule
        \textbf{Model} & \multicolumn{2}{c|}{\textbf{H1}} & \multicolumn{2}{c}{\textbf{H2}}\\
         & \textbf{P ($\uparrow$)} &  \multicolumn{1}{c|}{\textbf{S ($\uparrow$)}} & \textbf{P ($\uparrow$)}& \multicolumn{1}{c}{\textbf{S ($\uparrow$)}} \\ 
         \midrule
        Baseline (BLEU) & 0.32& 0.47 & 0.30 & 0.41 \\
        
        Tokenizer \& No Positional Embed. & 0.52 & 0.53 & 0.49 & 0.51 \\
        
        Positional Embed. \& No Tokenizer & 0.54 & 0.57 & 0.64 & 0.65 \\
        
        \textbf{Our full model} & 0.71& 0.75 & 0.69 & 0.70 \\
        \bottomrule
    \end{tabular}
   
    \label{tab:as}
\end{table}